\documentclass{article}
\usepackage[utf8]{inputenc}
\usepackage{authblk}

\title{Semi-supervised Conditional GANs}

\author[1]{Kumar Sricharan\thanks{sricharan.kumar@parc.com}}
\author[1]{Raja Bala}
\author[1]{Matthew Shreve}
\author[1]{\authorcr Hui Ding}
\author[2]{Kumar Saketh}
\author[1]{Jin Sun}
\affil[1]{\small{Interactive and Analytics Lab, Palo Alto Research Center, Palo Alto, CA}}
\affil[2]{\small{Verizon Labs, Palo Alto, CA}}

\date{\today}

\usepackage{natbib2}
\usepackage{graphicx}
\usepackage[a4paper, margin=1.5in]{geometry}
\usepackage{amsfonts}
\usepackage{hyperref}
\usepackage{subcaption}
\usepackage{wrapfig}

\begin{document}

\maketitle

\begin{abstract}
    We introduce a new model for building conditional generative models in a semi-supervised setting to conditionally generate data given attributes by adapting the GAN framework. The proposed semi-supervised GAN (SS-GAN) model uses a pair of stacked discriminators to  learn the marginal distribution of the data, and the conditional distribution of the attributes given the data respectively. In the semi-supervised setting, the marginal distribution (which is often harder to learn) is learned from the labeled + unlabeled data, and the conditional distribution is learned purely from the labeled data. Our experimental results demonstrate that this model performs significantly better compared to existing semi-supervised conditional GAN models. 
\end{abstract}

\section{Introduction}
Generative adversarial networks (GAN's)~\cite{goodfellow2014generative} are a recent popular technique for learning generative models for high-dimensional unstructured data (typically images). GAN's employ two networks - a generator G that is tasked with producing samples from the data distribution, and a discriminator D that aims to distinguish real samples from the samples produced by G. The two networks alternatively try to best each other, ultimately resulting in the generator G converging to the true data distribution. 

While most of the research on GAN's is focused on the unsupervised setting, where the data is comprised of unlabeled images, there has been research on conditional GAN's~\cite{gauthier2014conditional} where the goal is to learn a conditional model of the data, \emph{i.e.} to build a conditional model that can generate images given a particular attribute setting. In one approach~\cite{gauthier2014conditional}, both the generator and discriminator are fed attributes as side information so as to enable the generator to generate images conditioned on attributes. In an alternative approach proposed in ~\cite{odena2016conditional}, the authors build auxiliary classifier GAN's (AC-GAN's) where side information is reconstructed by the discriminator instead. Irrespective of the specific approach, this line of research focuses on the supervised setting where it is assumed that all the images have attribute tags.

Given that labels are expensive, it is of interest to explore semi-supervised settings where only a small fraction of the images have attribute tags, while a majority of the images are unlabeled. There has been some work on using GAN's in the semi-supervised setting. \cite{salimans2017improved} and \cite{springenberg2015unsupervised} use GAN's to perform semi-supervised classification by using a generator-discriminator pair to learn an unconditional model of the data and fine-tune the discriminator using the small amount of labeled data for prediction. However, we are not aware of work on building conditional models in the semi-supervised setting (see \ref{sec:framework} for details). The closest work we found was AC-GAN's, which can be extended to the semi-supervised setting in a straightforward manner (as was alluded to briefly by the authors in their paper). 

In the proposed semi-supervised GAN (SS-GAN) approach, we take a different route. We instead supply the side attribute information to the discriminator as is the case with supervised GAN's. We partition the discriminator's task of evaluating if the joint samples of images and attributes are real or fake into two separate tasks: (i) evaluating if the images are real or fake, and (ii) evaluating if the attributes given an image are real or fake. We subsequently use all the labeled and unlabeled data to assist the discriminator with the first task, and only the labeled images for the second task. The intuition behind this approach is that the marginal distribution of the images is much harder to model relative to the conditional distribution of the attributes given an image, and by separately evaluating the marginal and conditional samples, we can exploit the larger unlabeled pool to accurately estimate the marginal distribution.

Our main contributions in this work are as follows:
\begin{enumerate}
    \item We present the first extensive discussion of the semi-supervised conditional generation problem using GAN's.
    \item Related to (1), we apply the AC-GAN approach to the semi-supervised setting and present experimental results. 
    \item Finally, our main contribution is a new model called SS-GAN to effectively address the semi-supervised conditional generative modeling problem, which outperforms existing approaches including AC-GAN's for this problem. 
\end{enumerate}

\begin{figure}
\centering
\begin{subfigure}{.5\textwidth}
  \centering
  \includegraphics[width=.9\linewidth]{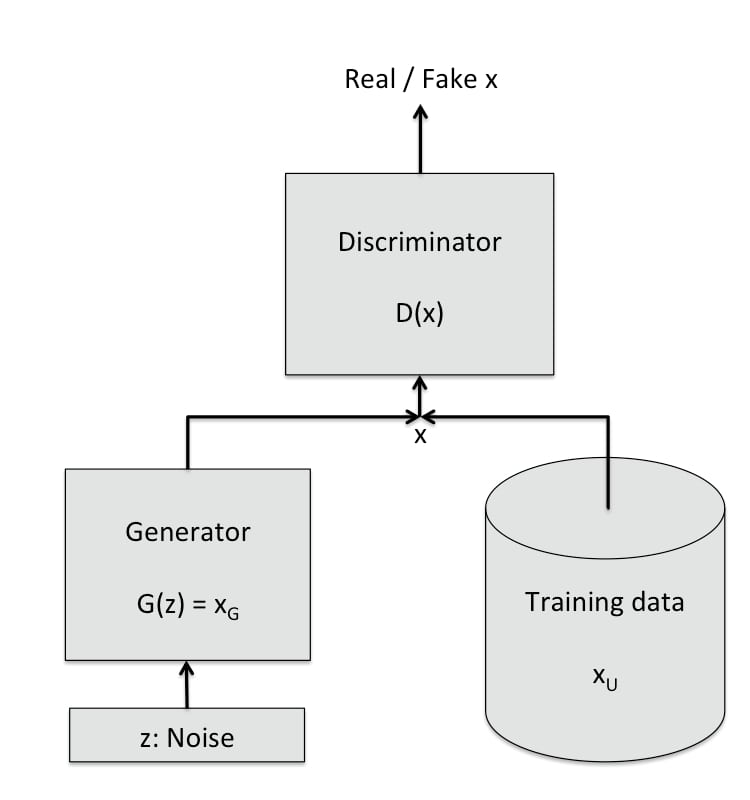}
  \caption{Unsupervised GAN}
  \label{fig:isub1}
\end{subfigure}%
\begin{subfigure}{.5\textwidth}
  \centering
  \includegraphics[width=.9\linewidth]{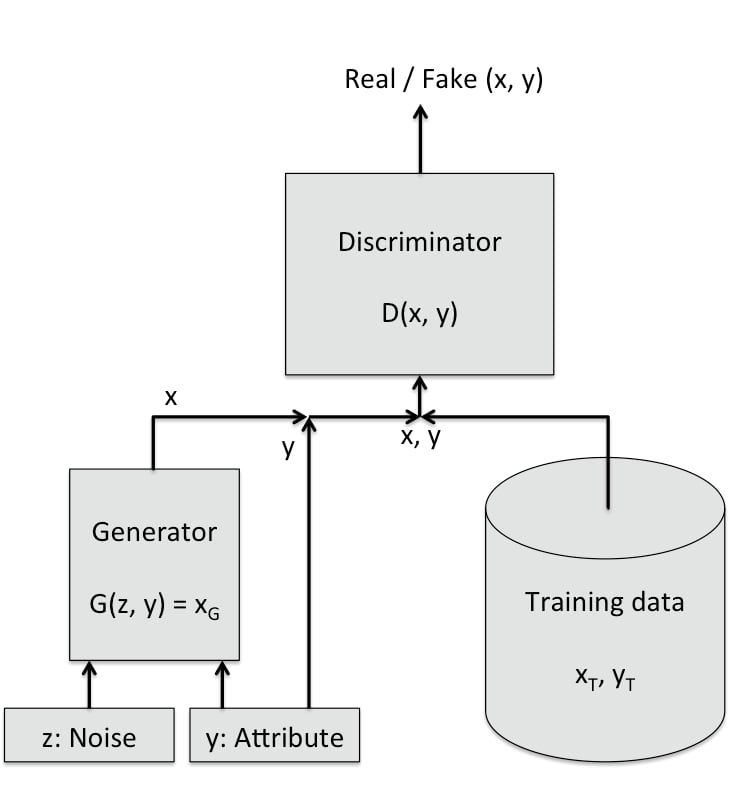}
  \caption{Conditional GAN}
  \label{fig:isub2}
\end{subfigure}

\bigskip
\begin{subfigure}{.5\textwidth}
  \centering
  \includegraphics[width=.9\linewidth]{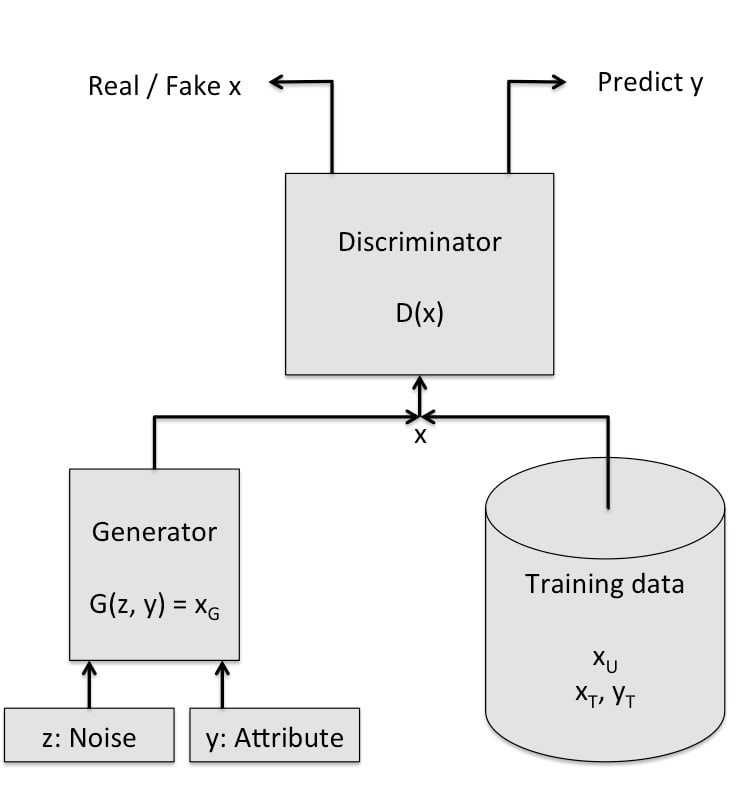}
  \caption{Auxiliary classifier GAN}
  \label{fig:isub3}
\end{subfigure}%
\begin{subfigure}{.5\textwidth}
  \centering
  \includegraphics[width=.9\linewidth]{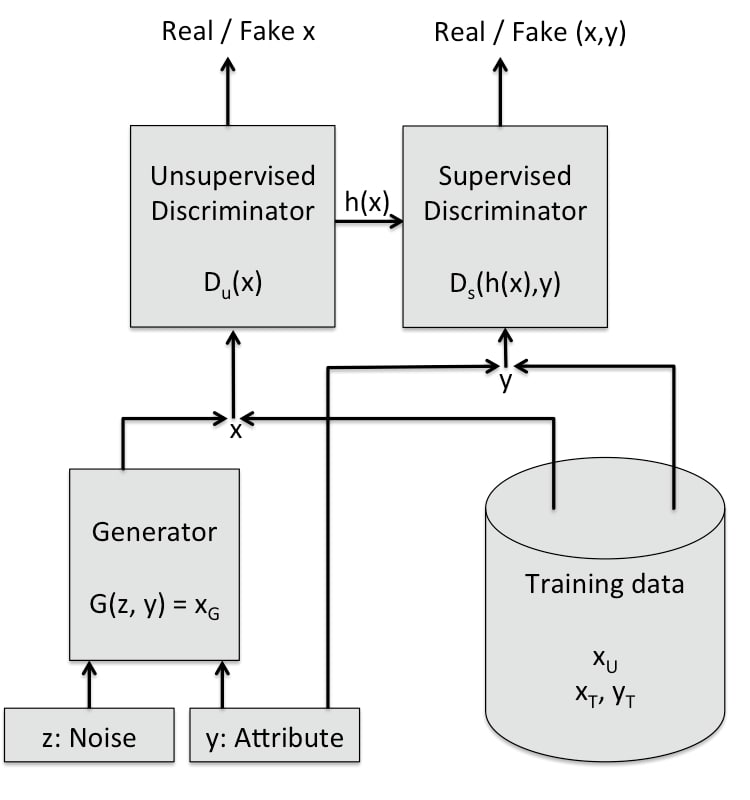}
  \caption{Semi-supervised stacked GAN}
  \label{fig:isub4}
\end{subfigure}
\caption{Illustration of 4 GAN models: (a) Unsupervised GAN, (b) Conditional GAN (for supervised setting), (c) Auxiliary Classifier GAN (for supervised and semi-supervised setting) and (d) proposed model semi-supervised GAN model. These models will be elaborated on further in the sequel.}
\label{fig:illustextra}
\end{figure}

The rest of this paper is organized as follows: In Section~\ref{sec:existing work}, we describe existing work on GAN's including details about the unsupervised, supervised and semi-supervised settings. Next, in Section~\ref{sec:new work}, we describe the proposed SS-GAN models, and contrast the model against existing semi-supervised GAN solutions. We present experimental results in Section~\ref{sec:exp}, and finally, we give our conclusions in Section~\ref{sec:conc}.

\section{Existing GAN's}
\label{sec:existing work}

\subsection{Framework}
\label{sec:framework}
We assume that our data-set is comprised of $n+m$ images $$\mathbb{X} = \{X_1, \ldots, X_n, X_{n+1}, \ldots, X_{n+m}\},$$ where the first $n$ images are accompanied by attributes $$\mathbb{Y} = \{Y_1, \ldots, Y_n\}.$$ Each $X_i$ is assumed to be of dimension $p_x \times p_y \times p_c$, where $p_c$ is the number of channels. The attribute tags $Y_i$ are assumed to be discrete variables of dimension $\{0,1,..,K-1\}^d$ - i.e., each attribute is assumed to be $d$-dimensional and each individual dimension of an attribute tag can belong to one of $K$ different classes. Observe that this accommodates class variables ($d=1$), and binary attributes ($K=2$). Finally, denote the joint distribution of images and attributes by $p(x,y)$, the marginal distribution of images by $p(x)$, and the conditional distribution of attributes given images by $p(y|x)$. Our goal is to learn a generative model $G(z,y)$ that can sample from $P(x|y)$ for a given $y$ by exploiting information from both the labeled and unlabeled sets. 

\subsection{Unsupervised GAN's}
\begin{figure}[!h]
\centering
\includegraphics[width=2in]{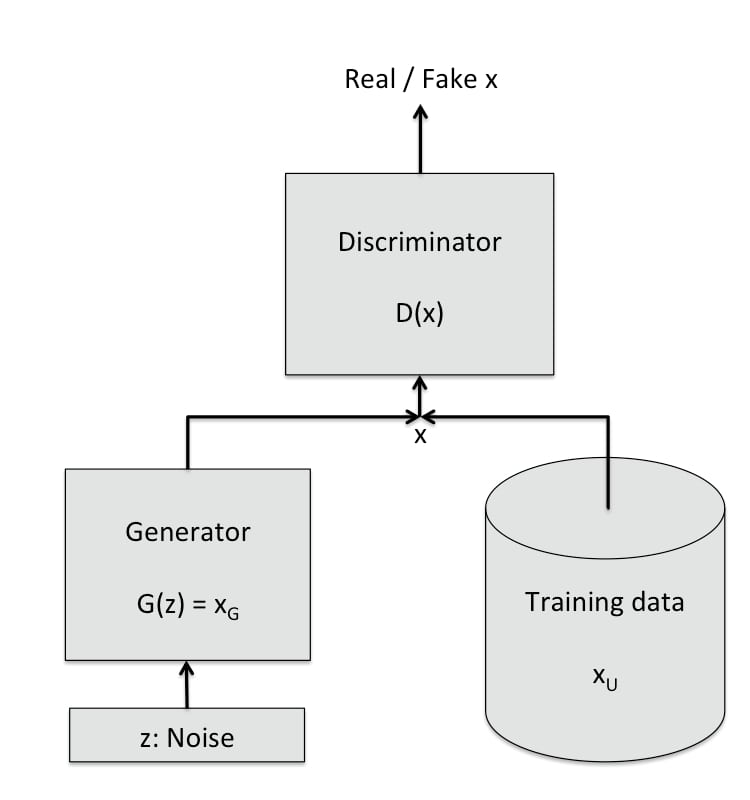}%
\caption{Illustration of unsupervised GAN model. }
\label{fig:unsupGAN}
\end{figure}
In the unsupervised setting $(n=0)$, the goal is to learn a generative model $G_u(z; \theta_u)$ that samples from the marginal image distribution $p(x)$, by transforming vectors of noise $z$ as $x=G_u(z; \theta_u)$. In order for $G_u()$ to learn this marginal distribution, a discriminator $D_u(x; \phi_u)$ is trained jointly~\cite{goodfellow2014generative}. The unsupervised loss functions for the generator and discriminator are as follows:
\begin{equation}
    \mathcal{L}_d^u(D_u,G_u) = \frac{1}{n+m} \left(\sum_{i=1}^{n+m} \log(D_u(X_i; \phi_u)) + \log(1-D_u(G_u(z_i; \theta_u); \phi_u))\right),
\end{equation}
and 
\begin{equation}
    \mathcal{L}_g^u(D_u,G_u) = \frac{1}{n+m} \left(\sum_{i=1}^{n+m} \log(D_u(G_u(z_i; \theta_u); \phi_u))\right),
\end{equation}
The above equations are alternatively optimized with respect to $\phi_u$ and $\theta_u$ respectively. The unsupervised GAN model is illustrated in ~\ref{fig:unsupGAN}.

\subsection{Supervised GAN's}
In the supervised setting (i.e., $m=0$), the goal is to learn a generative model $G_s(z, y; \theta_s)$ that samples from the conditional image distribution $p(x|y)$, by transforming vectors of noise $z$ as $x=G_s(z, y; \theta_s)$. There are two proposed approaches for solving this problem:

\subsubsection{Conditional GAN's}
\begin{figure}[!h]
\centering
\includegraphics[width=2.0in]{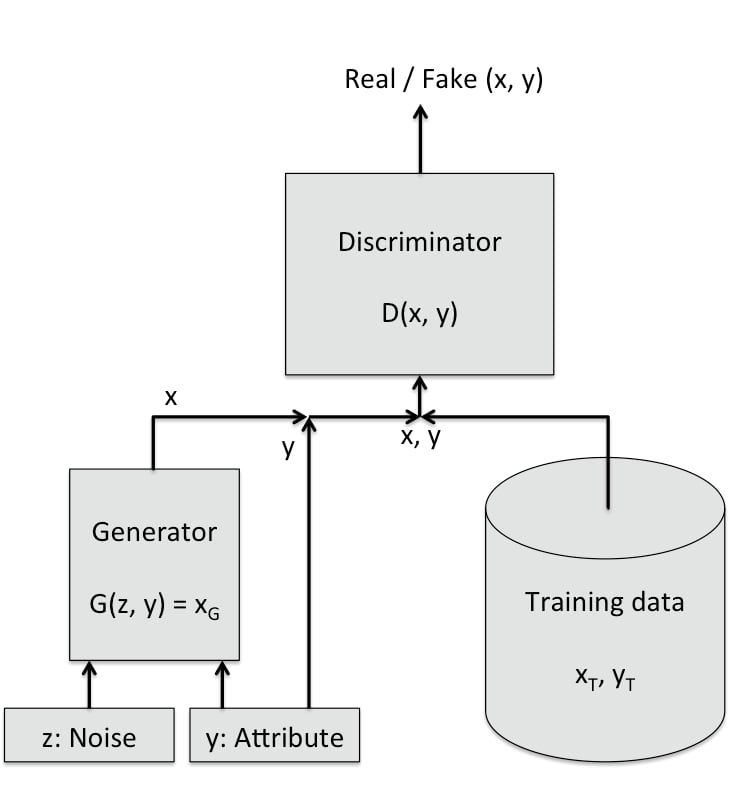}%
\caption{Illustration of Supervised Conditional GAN model. }
\label{fig:condGAN}
\end{figure}

In order for $G_s()$ to learn this conditional distribution, a discriminator $D_s(x, y; \phi_s)$ is trained jointly. The goal of the discriminator is to distinguish whether the joint samples $(x, y)$ are samples from the data or from the generator. The supervised loss functions for the generator and discriminator for conditional GAN (C-GAN) are as follows:
\begin{equation}
    \mathcal{L}_d^s(D_s,G_s) = \frac{1}{n} \left(\sum_{i=1}^{n} \log(D_s(X_i, Y_i; \phi_s)) + \log(1-D_s(G_s(z_i, Y_i; \theta_s), Y_i; \phi_s))\right),
\end{equation}
and 
\begin{equation}
    \mathcal{L}_g^s(D_s,G_s) = \frac{1}{n} \left(\sum_{i=1}^{n} \log(D_s(G_s(z_i, Y_i; \theta_s); \phi_s))\right),
\end{equation}
The above equations are alternatively optimized with respect to $\phi_s$ and $\theta_s$ respectively. The conditional GAN model is illustrated in ~\ref{fig:condGAN}.

\subsubsection{Auxiliary-classifier GAN's}
An alternative approach~\cite{odena2016conditional} to supervised conditional generation is to only supply the images $x$ to the discriminator, and ask the discriminator to additionally recover the true attribute information.  In particular, the discriminator $D_a(x; \phi_a)$ produces two outputs: (i) $D_{a(rf)}(x; \phi_a)$ and (ii) $D_{a(a)}(x, y; \phi_a)$, where the first output is the probability of $x$ being real or fake, and the second output is the estimated conditional probability of $y$ given $x$. In addition to the unsupervised loss functions, the generator and discriminator are jointly trained to recover the true attributes for any given images $X$. In particular, define the attribute loss function as 
\begin{equation}
    \mathcal{L}_a^a(D_{a(a)},G_a) = \frac{1}{n} \left(\sum_{i=1}^{n} \log(D_{a(a)}(X_i; Y_i \phi_a) + \log(D_{a(a)}(G_a(z_i, Y_i; \theta_a); Y_i \phi_a) \right).
\end{equation}
The loss function for the discriminator is given by 
\begin{equation}
    \mathcal{L}_d^a(D_a,G_a) = \mathcal{L}_d^u(D_{a(rf)},G_a) + \mathcal{L}_a^a(D_{a(a)},G_a), 
\end{equation}
and for the generator is given by 
\begin{equation}
    \mathcal{L}_g^a(D_a,G_a) = \mathcal{L}_g^u(D_{a(rf)},G_a) + \mathcal{L}_a^a(D_{a(a)},G_a),
\end{equation}

\subsubsection{Comparison between C-GAN and AC-GAN}
The key difference between C-GAN and AC-GAN is that instead of asking the discriminator to estimate the probability distribution of the attribute given the image as is the case in AC-GAN, C-GAN instead supplies discriminator $D_s$ with both $(x,y)$ and asks it to estimate the probability that $(x,y)$ is consistent with the true joint distribution $p(x,y)$.

While both models are designed to learn a conditional generative model, we did not find extensive comparisons between the two approaches in literature. To this end, we compared the performance of the two architectures using a suite of qualitative and quantitative experiments on a collection of data sets, and through our analysis (see Section~\ref{sec:exp}), determined that C-GAN typicaly outperforms AC-GAN in performance. 

\subsection{Semi-supervised GAN's}
We now consider the the semi-supervised setting where $m>0$, and typically $n<<m$. In this case, both C-GAN and AC-GAN can be applied to the problem. Because C-GAN required the attribute information to be fed to the discriminator, it can be applied only by trivially training it only on the labeled data, and throwing away the unlabeled data. We will call this model SC-GAN.

\begin{figure}[!h]
\centering
\includegraphics[width=2.0in]{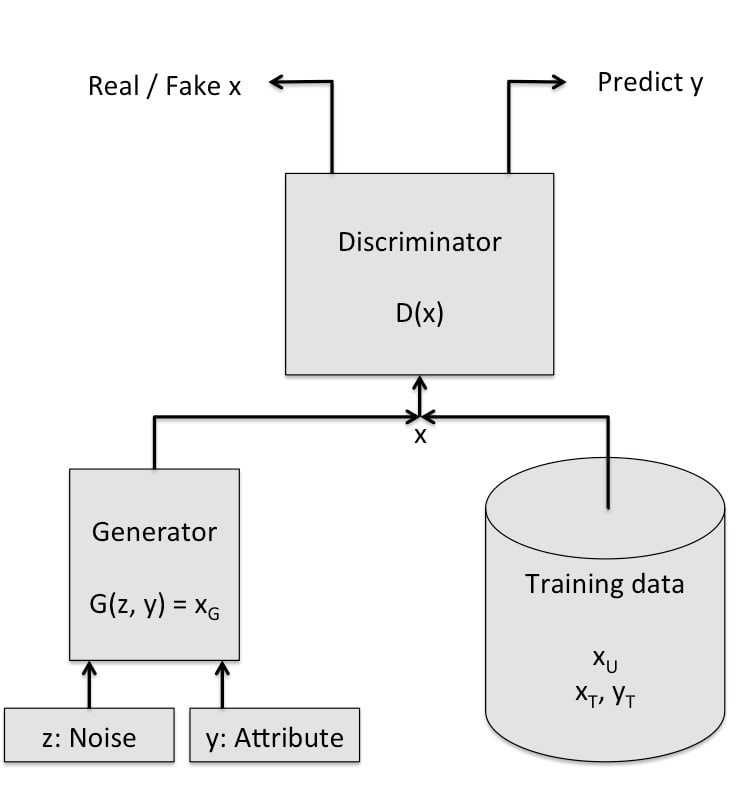}%
\caption{Illustration of Auxiliary Classifier GAN model. }
\label{fig:ACGAN}
\end{figure}

On the other hand, AC-GAN can be applied to this semi-supervised setting in a far more useful manner as alluded to by the authors in ~\cite{2017arXiv170403971X}. In particular, the adversarial loss terms $\mathcal{L}_d^u(D_a,G_a)$ and $\mathcal{L}_g^u(D_a,G_a)$ are evaluated over all the images in $\mathbb{X}$, while the attribute estimation loss term $\mathcal{L}_a^a(D_a,G_a)$ is evaluated over only the $n$ real images with attributes. We will call this model SA-GAN. This model is illustrated in ~\ref{fig:ACGAN}.

\section{Proposed Semi-supervised GAN}
\label{sec:new work}

We will now propose a new model for learning conditional generator models in a semi-supervised setting. This model aims to extend the C-GAN architecture to the semi-supervised setting that can exploit the unlabeled data unlike SC-GAN, by overcoming the difficulty of having to provide side information to the discriminator. By extending the C-GAN architecture, we aim to enjoy the same performance advantages over SA-GAN that C-GAN enjoys over AC-GAN. 

In particular, we consider a stacked discriminator architecture comprising of a pair of discriminators $D_u$ and $D_s$, with $D_u$ tasked with with distinguishing real and fake images $x$, and $D_s$ tasked with distinguishing real and fake (image, attribute) pairs $(x, y)$. Unlike in C-GAN, $D_u$ will separately estimate the probability that $x$ is real using both the labeled and unlabeled instances, and $D_s$ will separately estimate the probability that $y$ given $x$ is real using only the labeled instances. The intuition behind this approach is that the marginal distribution $p(x)$ is much harder to model relative to the conditional distribution $p(y|x)$, and by separately evaluating the marginal and conditional samples, we can exploit the larger unlabeled pool to accurately estimate the marginal distribution. 


\subsection{Model description}
Let $D_{ss}(x, y; \phi_{ss})$ denote the discriminator, which is comprised of two stacked discriminators: (i) $D_{s}(x; \phi_{ss})$ outputs the probability that the marginal image $x$ is real or fake, and (ii) $D_{u}(x, y; \phi_{ss})$ outputs the probability that the conditional attribute $y$ given the image $x$ is real or fake. The generator $G_{ss}(z, y; \theta_{ss})$ is identical to the generator in C-GAN and AC-GAN. The loss functions for the generator and the pair of discriminators are defined below:
\begin{equation}
    \mathcal{L}_d^{ss}(D_{u},G_{ss}) = \mathcal{L}_d^u(D_{u},G_{ss}),
\end{equation}
\begin{equation}
    \mathcal{L}_d^{ss}(D_{s},G_{ss}) = \mathcal{L}_d^s(D_{s},G_{ss}),
\end{equation}
and 
\begin{equation}
    \mathcal{L}_g^{ss}(D_{ss},G_{ss}) = \mathcal{L}_g^u(D_{ss(u)},G_{ss}) + \alpha\mathcal{L}_g^s(D_{ss(s)},G_{ss}),
\end{equation}
where $\alpha$ controls the effect of the conditional term relative to the unsupervised term.

\begin{figure}[!h]
\centering
\includegraphics[width=2.0in]{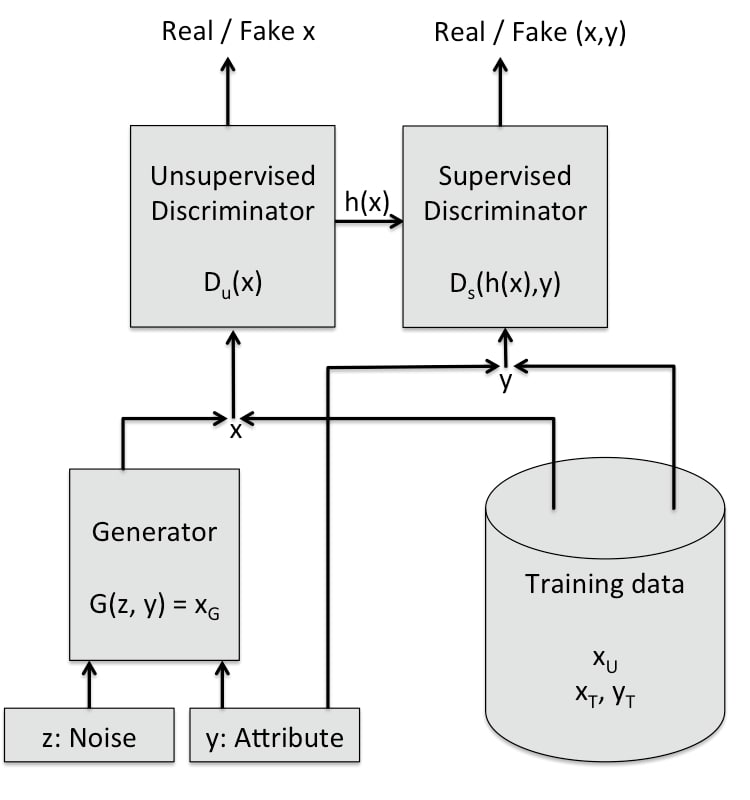}%
\caption{Illustration of proposed semi-supervised GAN model. Intermediate features $h(x)$ from the last but one layer of the unsupervised discriminator are concatenated with $y$ and fed to the supervised discriminator.}
\label{fig:SSGAN}
\end{figure}

\paragraph{Model architecture:} We design the model so that $D_{ss(u)}(x; \phi_{ss})$ depends only on the $x$ argument, and produces an intermediate output (last but one layer of unsupervised discriminator) $h(x)$, to which the argument $y$ is subsequently appended and fed to the supervised discriminator to produce the probability $D_{ss(s)}(x; \phi_{ss})$ that the joint samples $(x,y)$ are real/fake. The specific architecture is shown in Figure~\ref{fig:SSGAN}.

The advantage of this proposed model which supplies $x$ to $D_{ss(s)}$ via the features learned by $D_{ss(u)}$ over directly providing the $x$ argument to $D_{ss(s)}$ is that $D_{ss(s)}$ can not overfit to the few labeled examples, and instead must rely on the features general to the whole population in order to uncover the dependency between $x$ and $y$. 

For illustration, consider the problem of conditional face generation where one of the attributes of interest is eye-glasses. Also, assume that in the limited set of labeled images, only one style of eye-glasses (e.g., glasses with thick rims) are encountered. If so, then the conditional discriminator can learn features specific to rims to detect glasses if the entire image $x$ is available to the supervised discriminator. On the other hand, the features $h(x)$ learned by the unsupervised discriminator would have to generalize over all kinds of eyeglasses and not just rimmed eyeglasses specifically. In our stacked model, by restricting the supervised discriminator to access to the image $x$ through the features $h(x)$ learned by the unsupervised discriminator, we ensure that the supervised discriminator now generalizes to all different types of eyeglasses when assessing the conditional fit of the glasses attribute. 

\subsection{Convergence analysis of model}
Denote the distribution of the samples provided by the generator as $p'(x,y)$. Provided that the discriminator has sufficient modeling power, following Section 4.2 in \cite{goodfellow2014generative}, it follows that if we have sufficient data $m$, and if the discriminator is trained to convergence, $D_{u}(x; \phi_{ss})$ will converge to $p(x)/(p(x) + p'(x))$, and consequently, the generator will adapt its output so that $p'(x)$ will converge to $p(x)$. 

Because $n$ is finite and typically small, we are not similarly guaranteed that $D_{s}(x, y; \phi_{ss})$ will converge to $p(x,y)/(p(x,y) + p'(x,y))$, and that consequently, the generator will adapt its output so that $p'(x,y)$ will converge to $p(x,y)$. However, we make the key observation that because $p'(x)$ converges to $p(x)$ though the use of $D_u$, $D_{s}(x, y; \phi_{ss})$ will equivalently look to converge to $p(y|x)/(p(y|x) + p'(y|x))$, and given that these distributions are discrete, plus the fact that the supervised discriminator $D_{s}(x, y; \phi_{ss})$ operates on $x$ via the low-dimensional embedding $h(x)$, we hypothesize that $D_{s}(x, y; \phi_{ss})$ will successfully learn to closely approximate $p(y|x)/(p(y|x) + p'(y|x))$ even when $n$ is small. The joint use of $D_u$ and $D_s$ will therefore ensure that the joint distribution $p'(x,y)$ of the samples produced by the generator will converge to the true distribution $p(x,y)$. 

\section{Experimental results}
\label{sec:exp}
We propose a number of different experiments to illustrate the performance of the proposed SS-GAN over existing GAN approaches. 

\subsection{Models and datasets}
We compare the results of the proposed SS-GAN model against three other models:
\begin{enumerate}
    \item Standard GAN model applied to the full data-set (called C-GAN)
    \item Standard GAN model applied to only the labeled data-set (called SC-GAN)
    \item Supervised AC-GAN model applied to the full data-set (called AC-GAN)
    \item Semi-supervised AC-GAN model (called SA-GAN)
\end{enumerate}
We illustrate our results on 3 different datasets: (i) MNIST, (ii) celebA, and (iii) CIFAR10. 

\begin{figure}[!h]
\centering
\includegraphics[trim={0 0 0.25cm 0},clip,width=4.0in]{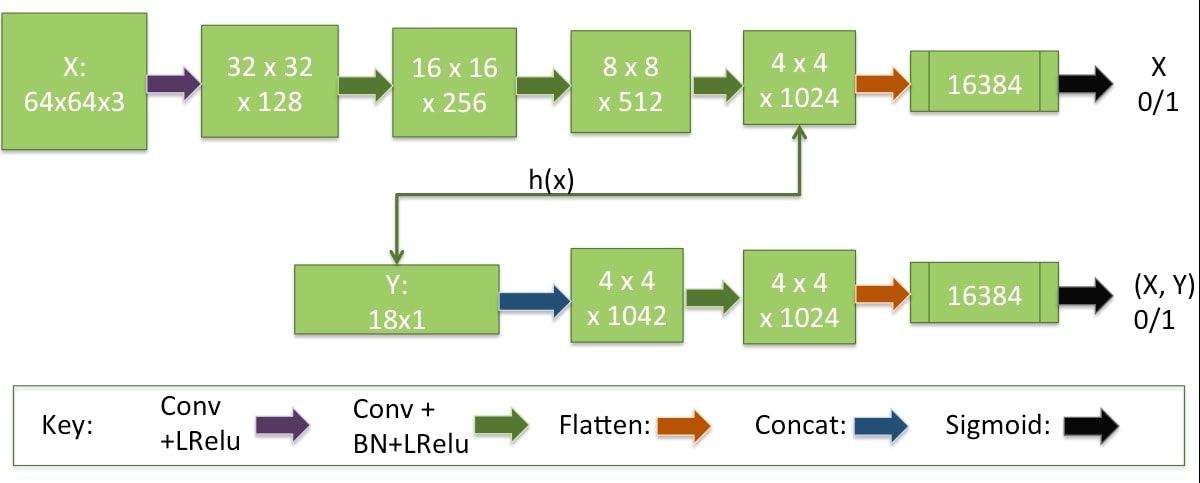}%
\caption{Illustration of SS-GAN discriminator for celebA dataset. The different layer operations in the neural network are illustrated by the different colored arrows (Conv = convolutional operator of stride 2, BN = Batch Normalization). }
\label{fig:dcgan_ss_arc}
\end{figure}

In all our experiments, we use the DCGAN architecture proposed in ~\cite{radford2015unsupervised}, with slight modifications to the generator and discriminator to accommodate the different variants described in the paper. These modifications primarily take the form of (i) concatenating the inputs $(x,y)$ and $(z,y)$ for the supervised generator and discriminator respectively, and adding an additional output layer to the discriminator in the case of AC-GAN, and connecting the last but one layer of $D_u$ to $D_s$ in the proposed SS-GAN. In particular, we use the same DCGAN architecture as in ~\cite{radford2015unsupervised} for MNIST and celebA, and a slightly modified version of the celebA architectures to accommodate the smaller 32x32 resolutions of the cifar10 dataset. The stacked DCGAN discriminator model for the celebA faces dataset is shown in Figure~\ref{fig:dcgan_ss_arc}. 

\subsection{Evaluation criteria}
We use a variety of different evaluation criteria to contrast SS-GAN against the models C-GAN, AC-GAN, SC-GAN and SA-GAN listed earlier. 
\begin{enumerate}
    \item Visual inspection of samples: We visually display a large collection of samples from each of the models and highlight differences in samples from the different models. 
    \item Reconstruction error: We optimize the inputs to the generator to reconstruct the original samples in the dataset (see Section 5.2 in ~\cite{2017arXiv170403971X}) with respect to squared reconstruction error. Given the drawbacks of reconstruction loss, we also compute the structural similarity metric (SSIM)~\cite{wang2004image} in addition to the reconstruction error.
    \item Attribute/class prediction from pre-trained classifier (for generator): We pre-train an attribute/class predictor from the entire training data set, and apply this predictor to the samples generated from the different models, and report the accuracy (RMSE for attribute prediction, 0-1 loss for class prediction). 
    \item Supervised learning error (for discriminator): We use the features from the discriminator and build classifiers on these features to predict attributes, and report the accuracy. 
    \item Sample diversity: To ensure that the samples being produced are representative of the entire population, and not just the labeled samples, we first train a classifier than can distinguish between the labeled samples (class label 0) and the unlabeled samples (class label 1). We then apply this classifier to the samples generated by each of the generators, and compute the mean probability of the samples belonging to class 0. The closer this number is to 0, the better the unlabeled samples are represented. 
\end{enumerate}

\subsection{MNIST}
The MNIST dataset contains 60,000 labeled images of digits. We perform semi-supervised training with a small randomly picked fraction
of these, considering setups with 10, 20, and 40 labeled examples. We ensure that each setup has a balanced number of examples
from each class. The remaining training images are provided without labels.

\subsubsection{Visual sample inspection}
\begin{figure}
\centering
\begin{subfigure}{.5\textwidth}
  \centering
  \includegraphics[width=.9\linewidth]{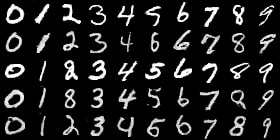}
  \caption{MNIST samples 1}
  \label{fig:sub1}
\end{subfigure}%
\begin{subfigure}{.5\textwidth}
  \centering
  \includegraphics[width=.9\linewidth]{figures/mnist_evaluate_samples_1.jpg}
  \caption{MNIST samples 2}
  \label{fig:sub2}
\end{subfigure}
\caption{2 sets of representative samples from the 5 models (each row from top to bottom corresponds to samples from C-GAN, AC-GAN, SC-GAN, SA-GAN and SS-GAN). SS-GAN's performance is close to the supervised models (C-GAN and AC-GAN). SA-GAN gets certain digit associations wrong, while SC-GAN generates copies of digits from the labeled set. }
\label{fig:mnistextra}
\end{figure}

\begin{figure}[!h]
\centering
\includegraphics[width=4.0in]{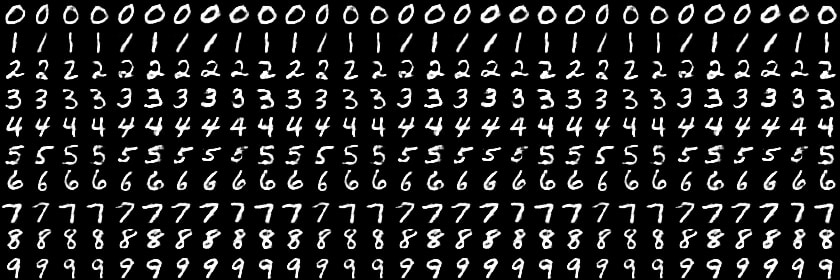}%
\caption{Digits generated by C-GAN model in the fully supervised setting (n=60000, m=0). }
\label{fig:mnistcase1}
\end{figure}

\begin{figure}[!h]
\centering
\includegraphics[width=4.0in]{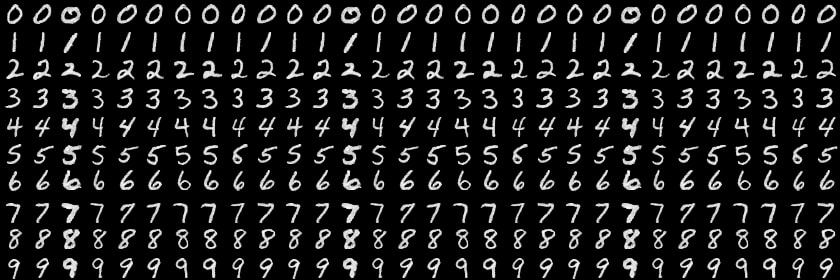}%
\caption{Digits generated by AC-GAN model in the fully supervised setting (n=60000, m=0). }
\label{fig:mnistcase0}
\end{figure}

\begin{figure}[!h]
\centering
\includegraphics[width=4.0in]{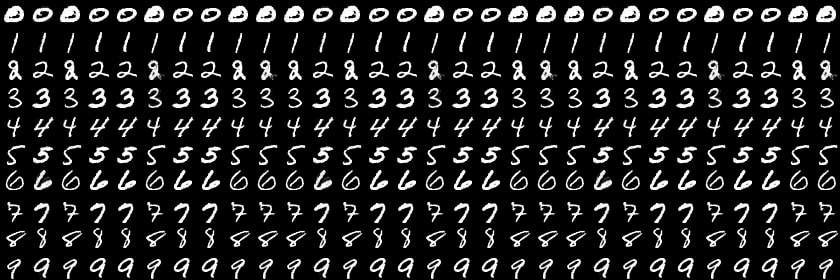}%
\caption{Digits generated by SC-GAN model in the small label supervised setting ) (n=20, m=0). }
\label{fig:mnistcase2}
\end{figure}

\begin{figure}[!h]
\centering
\includegraphics[width=4.0in]{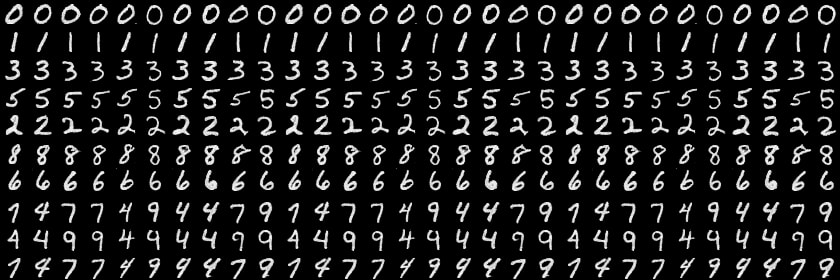}%
\caption{Digits generated by SA-GAN model in the semi-supervised setting (n=20, m=60000). }
\label{fig:mnistcase3}
\end{figure}

\begin{figure}[!h]
\centering
\includegraphics[width=4.0in]{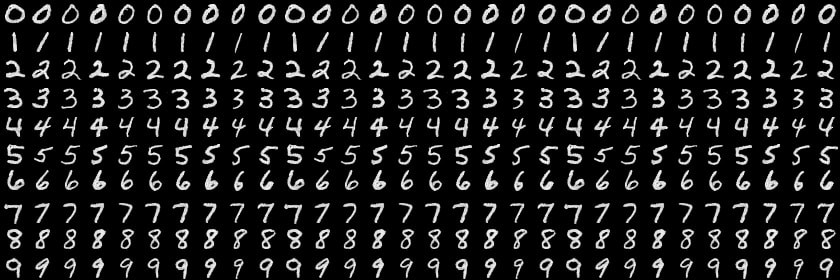}%
\caption{Digits generated by SS-GAN model in the semi-supervised setting (n=20, m=60000). SS-GAN samples are close to the quality achieved by the supervised C-GAN and AC-GAN models, avoids the incorrect attribute issues that affect the SA-GAN model, and the limited diversity of samples from SC-GAN.}
\label{fig:mnistcase4}
\end{figure}

In Figure~\ref{fig:mnistextra}, we show representative samples form the 5 different models for the case with $n=20$ labeled examples. In addition, in figures ~\ref{fig:mnistcase1}, \ref{fig:mnistcase0}, \ref{fig:mnistcase2}, \ref{fig:mnistcase3}, \ref{fig:mnistcase4}, we show more detailed results for this case with 20 labeled example (two examples per digit). In these detailed results, each row corresponds to a particular digit. Both C-GAN and AC-GAN successfully learn to model both the digits and the association between the digits and their class label. From the results, it is clear that SC-GAN learns to predict only the digit styles of each digit made available in the labeled set. While SA-GAN produces greater diversity of samples, it suffers in producing the correct digits for each label. SS-GAN on the other hand both produces diverse digits while also being accurate. In particular, its performance closely matches the performance of the fully supervised C-GAN and AC-GAN models. This is additionally borne out by the quantitative results shown in Tables ~\ref{tab:mnist10}, \ref{tab:mnist20} and \ref{tab:mnist40} for the cases $n=10, 20$ and $40$ respectively, as shown below. 

\begin{table}[!h]
    \centering
    \begin{small}
    \begin{tabular}{c|c|c|c|c}
    Samples source &  Class pred. error & Recon. error & Sample diversity & Discrim. error\\ \hline 
    True samples & 0.0327 & N/A & 0.992 & N/A\\
    Fake samples & N/A & N/A & 1.14e-05 & N/A\\
    C-GAN & 0.0153 & 0.0144 & 1.42e-06 & 0.1015\\
    AC-GAN & 0.0380 & 0.0149 & 1.49e-06 & 0.1140 \\
    SC-GAN & 0.0001 & 0.1084 & 0.999 & 0.095 \\
    SA-GAN & 0.3091 & 0.0308 & 8.62e-06 & 0.1062 \\
    SS-GAN & 0.1084 & 0.0320 & 0.0833 & 0.1024 \\
    \end{tabular}
    \end{small}
    \caption{Compilation of quantitative results for the MNIST dataset for $n=10$. }
    \label{tab:mnist10}
\end{table}

\begin{table}[!h]
    \centering
    \begin{small}
    \begin{tabular}{c|c|c|c|c}
    Samples source &  Class pred. error & Recon. error & Sample diversity & Discrim. error\\ \hline 
    True samples & 0.0390 & N/A & 0.994 & N/A\\
    Fake samples & N/A & N/A & 2.86e-05 & N/A\\
    C-GAN & 0.0148 & 0.01289 & 8.74e-06 & 0.1031\\
    AC-GAN & 0.0189 & 0.01398 & 9.10e-06 & 0.1031 \\
    SC-GAN & 0.0131 & 0.0889 & 0.998 & 0.1080 \\
    SA-GAN & 0.2398 & 0.02487 & 2.18e-05 & 0.1010 \\
    SS-GAN & 0.1044 & 0.0160 & 2.14e-05 & 0.1014 \\
    \end{tabular}
    \end{small}
    \caption{Compilation of quantitative results for the MNIST dataset for $n=20$. }
    \label{tab:mnist20}
\end{table}

\begin{table}[!h]
    \centering
    \begin{small}
    \begin{tabular}{c|c|c|c|c}
    Samples source &  Class pred. error & Recon. error & Sample diversity & Discrim. error\\ \hline 
    True samples & 0.0390 & N/A & 0.993 & N/A\\
    Fake samples & N/A & N/A & 1.63e-05 & N/A\\
    C-GAN & 0.0186 & 0.0131 & 1.36e-05 & 0.1023\\
    AC-GAN & 0.0141 & 0.0139 & 6.84e-06 & 0.1054 \\
    SC-GAN & 0.0228 & 0.080 & 0.976 & 0.1100 \\
    SA-GAN & 0.1141 & 0.00175 & 1.389e-05 & 0.1076 \\
    SS-GAN & 0.0492 & 0.0135 & 3.54e-05 & 0.1054 \\
    \end{tabular}
    \end{small}
    \caption{Compilation of quantitative results for the MNIST dataset for $n=40$. }
    \label{tab:mnist40}
\end{table}

\subsubsection{Discussion of quantitative results}
The fraction of incorrectly classified points for each source, the reconstruction error, the sample diversity metric and the discriminator error is shown in Tables~\ref{tab:mnist10}, \ref{tab:mnist20} and \ref{tab:mnist40} below. SS-GAN comfortably outperforms SA-GAN with respect to classification accuracy, and comfortably beats SC-GAN with respect to reconstruction error (due to the limited sample diversity of SC-GAN). The sample diversity metric for SS-GAN is slightly worse compared to SA-GAN, but significantly better than SC-GAN. Taken together, in conjunction with the visual analysis of the samples, these results conclusively demonstrate that SS-GAN is superior to SA-GAN and SC-GAN in the semi-supervised setting. 

From the three sets of results for the different labeled sample sizes ($n=10, 20$ and $40$), we see that the performance of all the models increases smoothly with increasing sample size, but with SSGAN still outperforming the other two semi-supervised models for each of the settings for the number of labeled samples. 

\subsubsection{Semi-supervised learning error}
For MNIST, we run an additional experiment, where  we draw samples from the various generators, train a classifier using each set of samples, and record the test error performance of this classifier. On MNIST, with 20 labeled examples, we show the accuracy of classifiers trained using samples generated from different models using MNIST in Table~\ref{tab:mnist3}. 
\begin{table}[!h]
    \centering
    \begin{tabular}{c|c}
    Samples source    &  10-fold 0-1 error \\ \hline 
    C-GAN & 5.1\\
    AC-GAN & 5.2\\
    SC-GAN & 12.9 \\
    SA-GAN & 24.3\\
    SS-GAN & 5.4\\
    \end{tabular}
    \caption{Classifier accuracy using samples generated from different models for MNIST.}
    \label{tab:mnist3}
\end{table}
From the results in table~\ref{tab:mnist3}, we see that our model SS-GAN is performing close to the supervised models. In particular, we note that these results are the state-of-the-art for MNIST given just 20 labeled examples (please see ~\cite{salimans2017improved} for comparison). However, the performance as the number of labeled examples increases remains fairly stationary, and furthermore is not very effective for more complex datasets such as CIFAR10 and celebA, indicating that this approach of using samples from GAN's to train classifiers should be restricted to very low sample settings for simpler data sets like MNIST.

\subsection{celebA dataset results}
CelebFaces Attributes Dataset (CelebA)~\cite{liu2015faceattributes} is a large-scale face attributes dataset with more than 200K celebrity images, each with 40 attribute annotations. The images in this dataset cover large pose variations and background clutter. Of the 40 attributes, we sub-select the following 18 attributes: 0: 'Bald', 1: 'Bangs', 2: 'Black Hair', 3: 'Blond Hair', 4: 'Brown Hair', 5: 'Bushy Eyebrows', 6: 'Eyeglasses', 7: 'Gray Hair', 8: 'Heavy Makeup', 9: 'Male’, 10: 'Mouth Slightly Open', 11: 'Mustache', 12: 'Pale Skin', 13: 'Receding Hairline', 14: 'Smiling', 15: 'Straight Hair’, 16: 'Wavy Hair’,  17:’Wearing Hat’. 

\subsubsection{Visual sample inspection} 
\begin{figure}
\centering
\begin{subfigure}{1\textwidth}
  \centering
  \includegraphics[width=.9\linewidth]{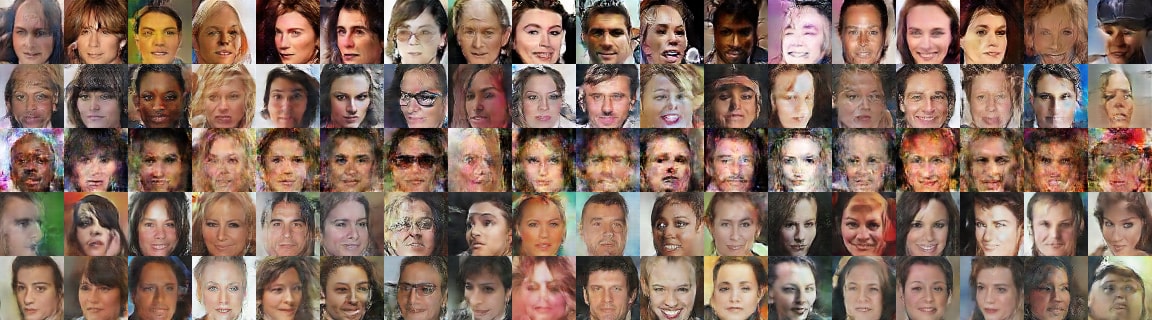}
  \caption{CelebA samples 1}
  \label{fig:csub1}
\end{subfigure}%

\bigskip
\begin{subfigure}{1\textwidth}
  \centering
  \includegraphics[width=.9\linewidth]{figures/celeba_evaluate_samples_1.jpg}
  \caption{CelebA samples 2}
  \label{fig:csub2}
\end{subfigure}
\caption{2 sets of representative samples from the 5 models (each row from top to bottom corresponds to samples from C-GAN, AC-GAN, SC-GAN, SA-GAN and SS-GAN respectively). SS-GAN's performance is close to the supervised models (C-GAN and AC-GAN). SA-GAN gets certain associations wrong (e.g., attributes 0, 7 and 12), while SC-GAN produces samples of poor visual quality. }
\label{fig:celebaextra}
\end{figure}

\begin{figure}[!h]
\centering
\includegraphics[width=4.0in]{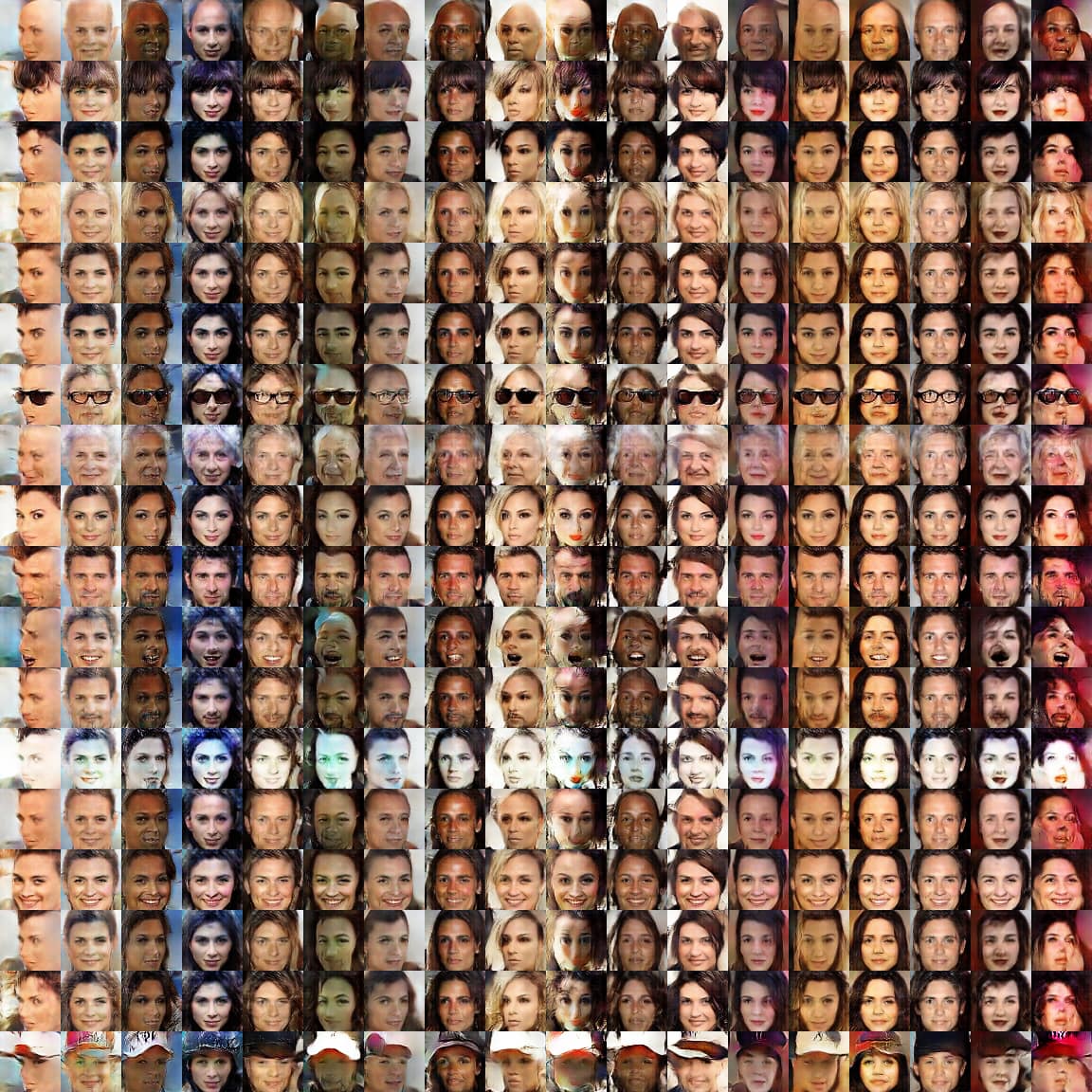}%
\caption{Faces generated by C-GAN model in the fully supervised setting (n=51200, m=0). }
\label{fig:case0}
\end{figure}

\begin{figure}[!h]
\centering
\includegraphics[width=4.0in]{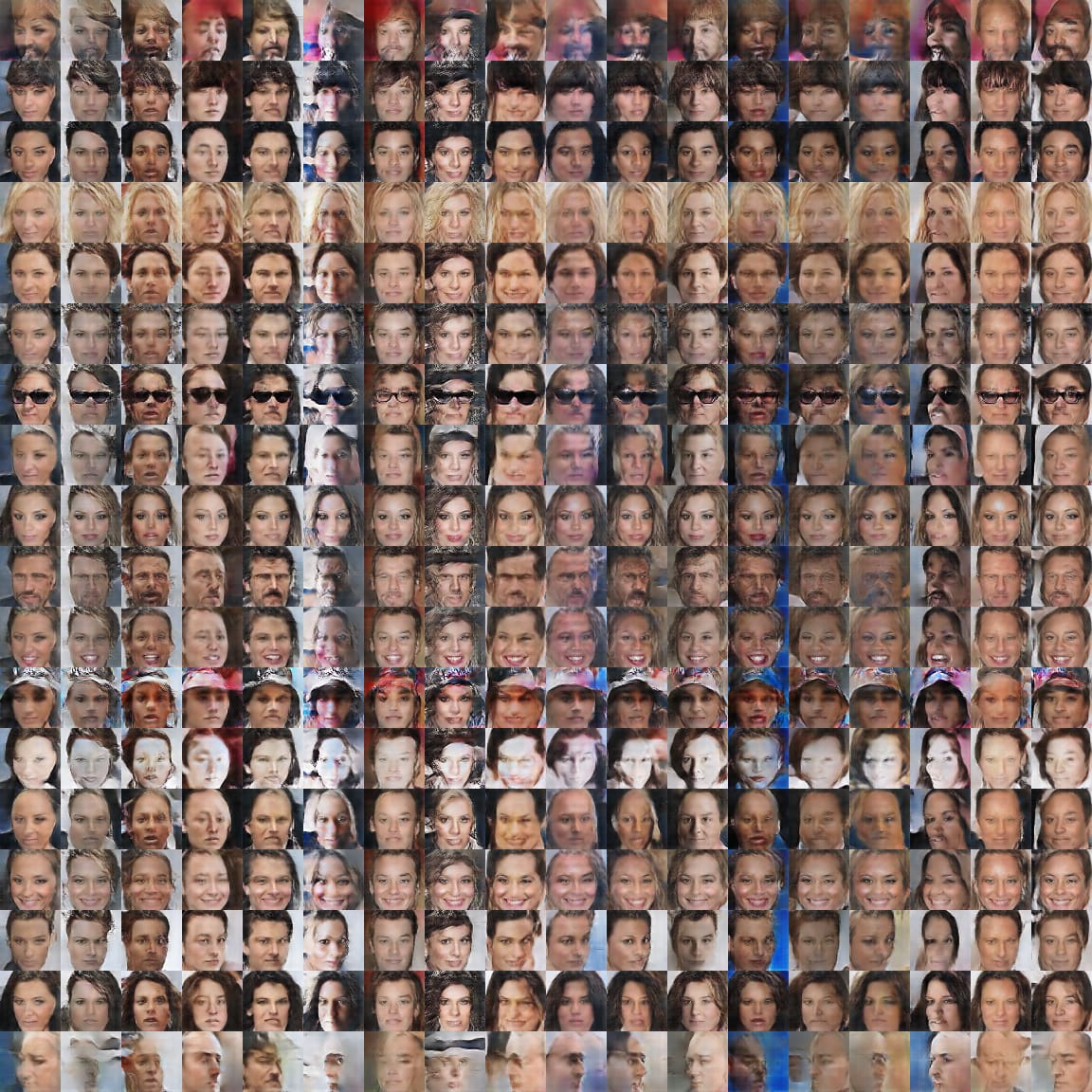}%
\caption{Faces generated by AC-GAN model in the fully supervised setting (n=51200, m=0). }
\label{fig:case1}
\end{figure}

\begin{figure}[!h]
\centering
\includegraphics[width=4.0in]{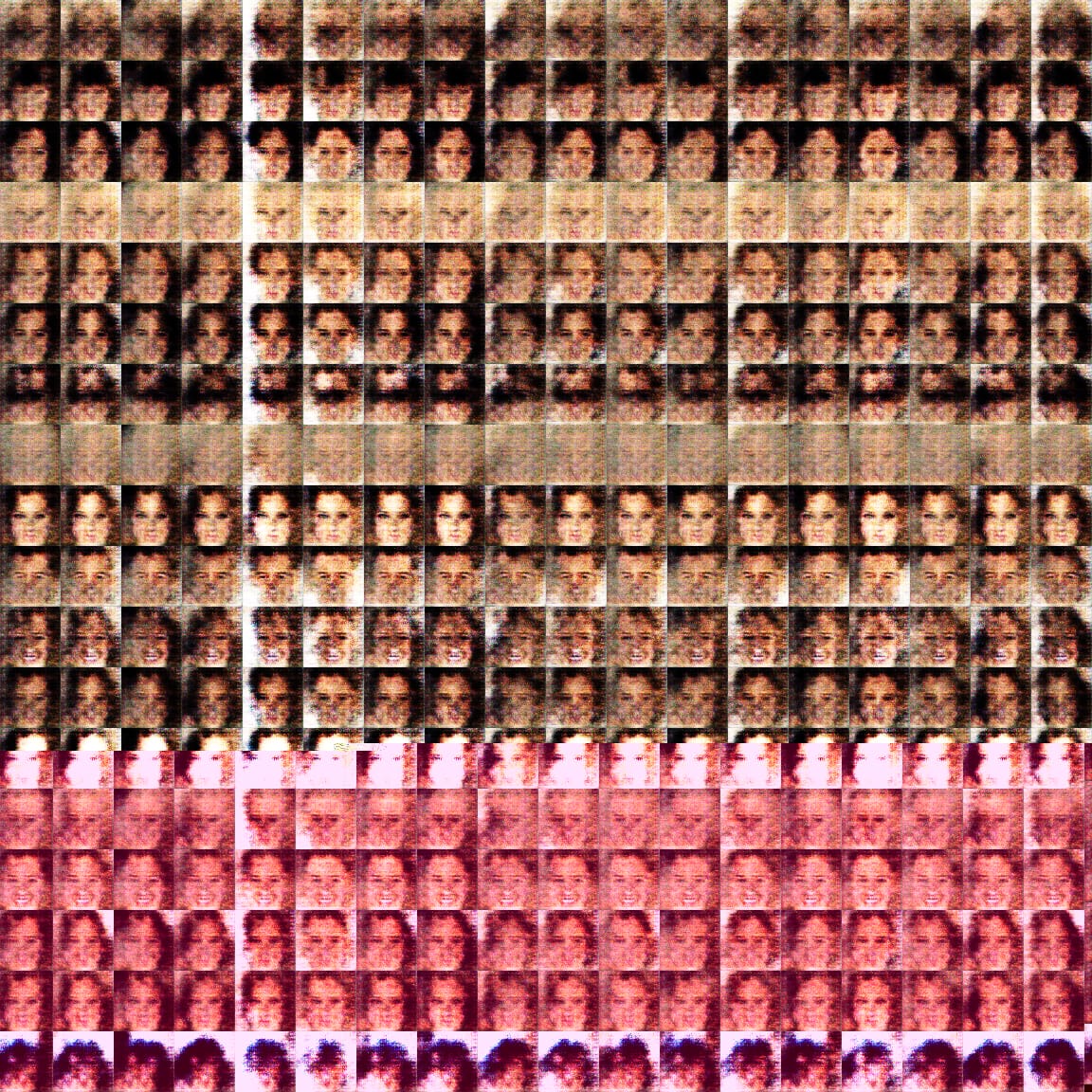}%
\caption{Faces generated by SC-GAN model in the small label supervised setting  (n=1024, m=0). }
\label{fig:case2}
\end{figure}

\begin{figure}[!h]
\centering
\includegraphics[width=4.0in]{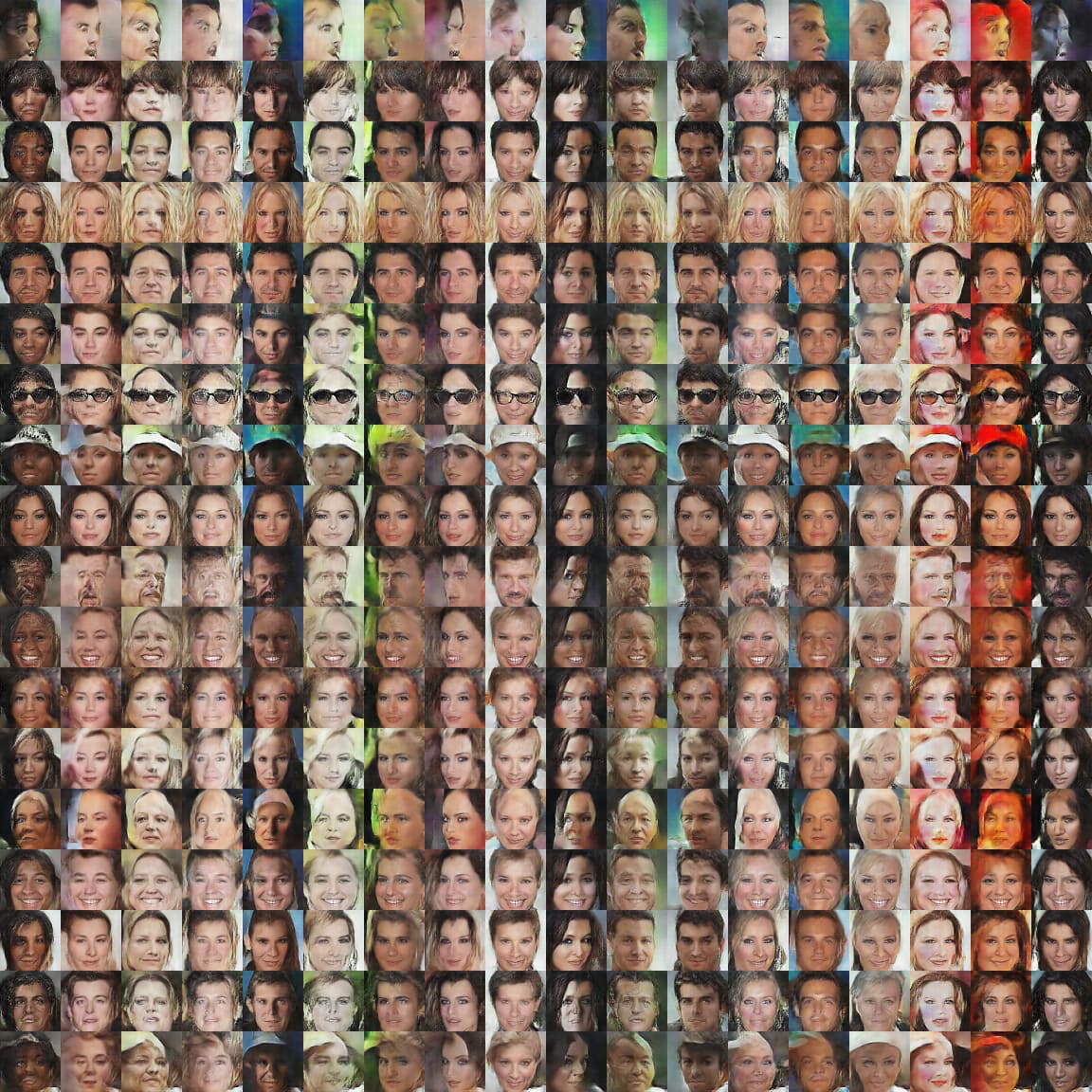}%
\caption{Faces generated by SA-GAN model in the semi-supervised setting (n=1024, m=50000). }
\label{fig:case3}
\end{figure}

\begin{figure}[!h]
\centering
\includegraphics[width=4.0in]{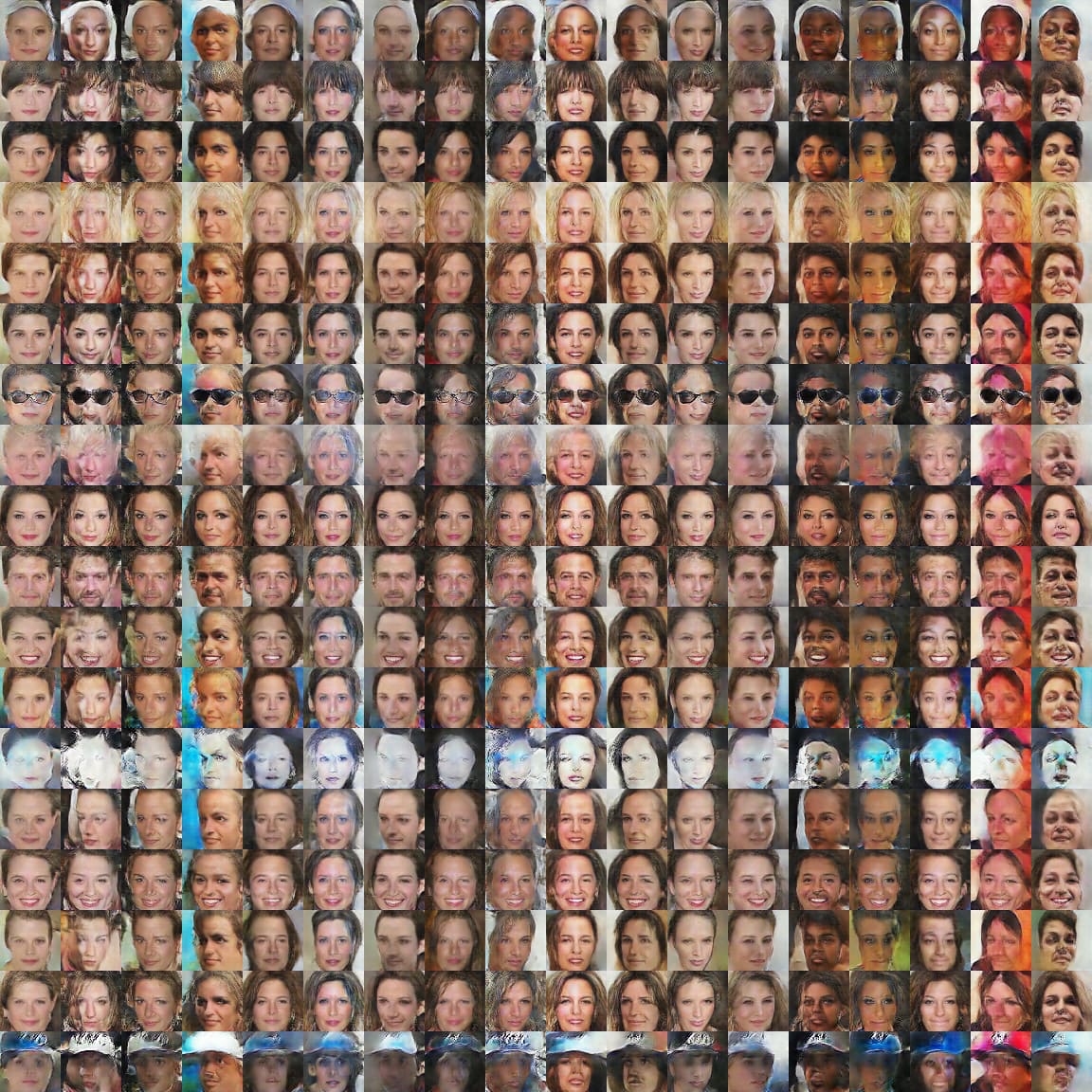}%
\caption{Faces generated by SS-GAN model in the semi-supervised setting (n=1024, m=50000). SS-GAN samples are close to the quality achieved by the supervised C-GAN and AC-GAN models, avoids the incorrect attribute issues that affect the SA-GAN model, and the poor quality of samples from SC-GAN.}
\label{fig:case4}
\end{figure}

In Figure~\ref{fig:celebaextra}, we show representative samples form the 5 different models for the case with $n=1000$ labeled examples for the celebA dataset. Each row correponds to an individual model, and each column corresponds to one of the 18 different attributes listed above. In addition, we show more detailed samples generated by the 5 different models in figures ~\ref{fig:case1}, \ref{fig:case0}, \ref{fig:case2}, \ref{fig:case3}, and  \ref{fig:case4}. In each of these figures, each row corresponds to a particular attribute type while all the other attributes are set to 0. From the generated samples, we can once again see that the visual samples produced by SS-GAN are close to the quality of the samples generated by the fully supervised models C-GAN and AC-GAN. SC-GAN when applied to the subset of data produces very poor results (significant mode collapse + poor quality of the generated images), while SA-GAN is relatively worse when compared to SS-GAN. For instance, SA-GAN produces images with incorrect attributes for attributes 0 (faces turned to a side instead of bald), 7 (faces with hats instead of gray hair), and 12 (generic faces instead of faces with pale skin). 

\begin{table}[!h]
    \centering
    \begin{small}
    \begin{tabular}{c|c|c|c|c|c}
    Samples source  &  Attribute RMSE & Recon. error & SSIM & Sample diversity & Disc. error \\ \hline 
    True samples & 0.04 & N/A & N/A & 0.99 & N/A\\
    Fake samples & N/A & N/A & N/A & 0.001 & N/A\\
    C-GAN & 0.25 & 0.036 & 0.497 & 0.002 & 0.07\\
    AC-GAN & 0.29 & 0.047 & 0.076 & 0.005 & 0.06\\
    SC-GAN & 0.26 & 0.343 & 0.143 & 0.454 & 0.01\\
    SA-GAN & 0.36 & 0.042 & 0.167 & 0.006 & 0.07\\
    SS-GAN & 0.31 & 0.040 & 0.217 & 0.004 & 0.03\\
    \end{tabular}
    \end{small}
    \caption{Compilation of quantitative results for the celebA dataset. Across the joint set of metrics, SS-GAN achieves performance close to the supervised C-GAN and AC-GAN models, while performing much better than either of the semi-supervised models - SC-GAN and SA-GAN.}
    \label{tab:faces1}
\end{table}

\subsubsection{Discussion of quantitative results}
The four different quantitative metrics - The attribute prediction error, the reconstruction error, the sample diversity metric and the discriminator error - are shown in Table~\ref{tab:faces1}. 

SS-GAN comfortably outperforms SA-GAN and achieves results close to the fully supervised models for the attribute prediction error metric. It is interesting to note that SC-GAN produces better attribute prediction error numbers than the SA-GAN model, while producing notably worse samples. 
We also find that with respect to reconstruction error and the SSIM metric, SS-GAN marginally out performs SA-GAN while coming close to the performance of the supervised C-GAN and AC-GAN models. As expected, SC-GAN performs poorly in this case. We also find that SS-GAN has a fairly low sample diversity score, marginally higher than C-GAN, but better than SA-GAN, and better even than the fully supervised AC-GAN.
Finally, SS-GAN comfortably outperforms SA-GAN and achieves results close to the fully supervised model with respect to the discriminator feature error. 

\subsection{cifar10 dataset}
The CIFAR-10 dataset~\cite{krizhevsky2009learning} consists of 60000 32x32 colour images in 10 classes, with 6000 images per class. There are 50000 training images and 10000 test images. The following are the 10 classes: airplane, automobile, bird, cat, deer, dog, frog, horse, ship, truck. 

\subsubsection{Visual sample inspection} 
\begin{figure}[!h]
\centering
\includegraphics[width=4.0in]{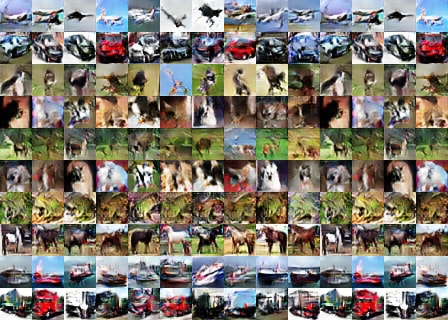}%
\caption{Cifar10 images generated by C-GAN model in the fully supervised setting (n=50000, m=0). }
\label{fig:cifar101}
\end{figure}

\begin{figure}[!h]
\centering
\includegraphics[width=4.0in]{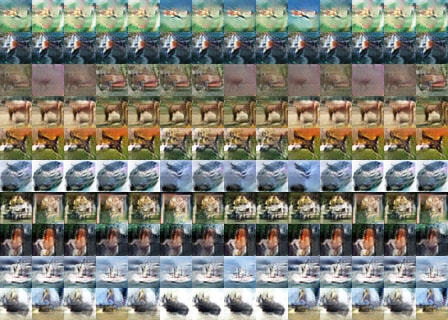}%
\caption{Cifar10 images generated by AC-GAN model in the fully supervised setting (n=50000, m=0). }
\label{fig:cifar100}
\end{figure}

\begin{figure}[!h]
\centering
\includegraphics[width=4.0in]{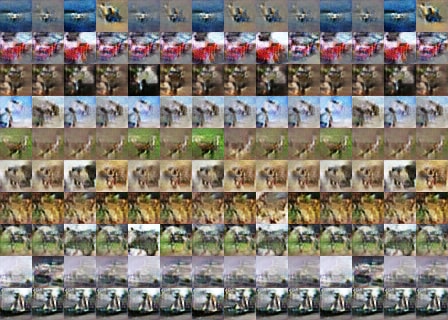}%
\caption{Cifar10 images generated by SC-GAN model in the semi-supervised setting (n=4000, m=50000). }
\label{fig:cifar102}
\end{figure}

\begin{figure}[!h]
\centering
\includegraphics[width=4.0in]{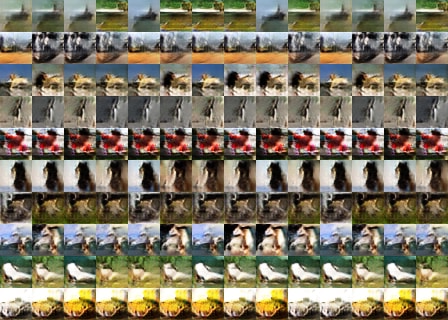}%
\caption{Cifar10 images generated by SA-GAN model in the semi-supervised setting (n=4000, m=50000). }
\label{fig:cifar103}
\end{figure}

\begin{figure}[!h]
\centering
\includegraphics[width=4.0in]{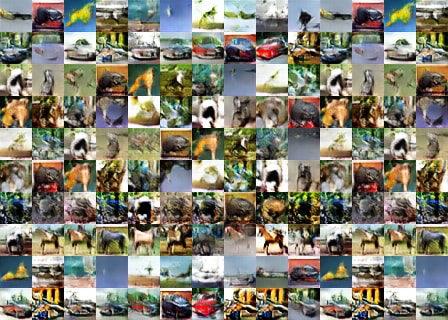}%
\caption{Cifar10 images generated by SS-GAN model in the semi-supervised setting (n=4000, m=50000). SS-GAN comes close to C-GAN with respect to quality of the samples, and avoids the mode collapse problems of AC-GAN, SA-GAN and SC-GAN.  }
\label{fig:cifar104}
\end{figure}

From the generated samples in figures~\ref{fig:cifar101}, \ref{fig:cifar100}, \ref{fig:cifar102}, \ref{fig:cifar103} and \ref{fig:cifar104}, we can see that the visual samples produced by SS-GAN are close to the quality of the samples generated by C-GAN. All the other three models, AC-GAN, SA-GAN, and SC-GAN suffer from significant mode collapse. We especially found the poor results of AC-GAN in the fully supervised case surprising, especially given the good performance of C-GAN on cifar10, and the good performance of AC-GAN on the MNIST and celebA datasets. 

\begin{table}[!h]
    \centering
    \begin{small}
    \begin{tabular}{c|c|c|c|c|c}
    Samples source  &  Class pred. error & Recon. error & SSIM & Sample diversity & Disc. error \\ \hline 
    True samples & 0.098 & N/A & N/A & 1.00 & N/A\\
    Fake samples & N/A & N/A & N/A & 1.21e-07 & N/A\\
    C-GAN & 0.198 & 0.041 & 0.501 & 1.39e-07 & 0.874\\
    AC-GAN & 0.391 & 0.204 & 0.024 & 1.41e-06 & 0.872\\
    SC-GAN & 0.355 & 0.213 & 0.026 & 0.999 & 0.870\\
    SA-GAN & 0.0.468 & 0.173 & 0.021 & 2.30e-06 & 0.874\\
    SS-GAN & 0.299 & 0.061 & 0.042 & 6.54e-06 & 0.891\\
    \end{tabular}
    \end{small}
    \caption{Compilation of quantitative results for the cifar10 dataset. Across the joint set of metrics, SS-GAN achieves performance close to the supervised C-GAN and AC-GAN models, while performing much better than either of the semi-supervised models - SC-GAN and SA-GAN. }
    \label{tab:cifar10}
\end{table}

\subsubsection{Discussion of quantitative results}
The different quantitative metrics computed against the cifar10 datasets are shown in Table~\ref{tab:cifar10}. In our experiments, we find that the samples generated by SS-GAN are correctly classified 70 percent of the time, which is second best after C-GAN and is off from the true samples by 15 percent. We also find that the reconstruction error for SS-GAN comes close to the performance of C-GAN and comfortably out performs the other three models. This result is consistent with the visual inspection of the samples. The sample diversity metric for SS-GAN is significantly better than SC-GAN, and comparable to the other three models. 

\section{Conclusion and discussion}
\label{sec:conc}
We proposed a new GAN based framework for learning conditional models in a semi-supervised setting. Compared to the only existing semi-supervised GAN approaches (i.e., SC-GAN and SA-GAN), our approach shows a marked improvement in performance over several datasets including MNIST, celebA and CIFAR10 with respect to visual quality of the samples as well as several other quantitative metrics. In addition, the proposed technique comes with theoretical convergence properties even in the semi-supervised case where the number of labeled samples $n$ is finite. 

From our results on all three of these datasets, we can conclude that the proposed SS-GAN performs almost as well as the fully supervised C-GAN and AC-GAN models, even when provided with very low number of labeled samples (down to the extreme limit of just one sample per class in the case of MNIST). In particular, it comfortably outperforms the semi-supervised variants of C-GAN and AC-GAN (SC-GAN and SA-GAN respectively). While the superior performance over SC-GAN is clearly explained by the fact that SC-GAN is only trained on the labeled data set, the performance advantage of SS-GAN over SA-GAN is not readily apparent. We explicitly discuss the reasons for this below:

\subsection{Why does SS-GAN work better than SA-GAN?}
\begin{enumerate}
    \item Unlike AC-GAN where the discriminator is tasked with recovering the attributes, in C-GAN, the discriminator is asked to estimate if the pair $(x,y)$ is real or fake. This use of adversarial loss that classifies $(x,y)$ pairs as real or fake over the cross-entropy loss that asks the discriminator to recover $y$ from $x$ seems to work far better as demonstrated by our experimental results. 
    Our proposed SS-GAN model learns the association between $x$ and $y$ using an adversarial loss as is the case with C-GAN, while SA-GAN uses the cross-entropy loss over the labeled samples.
    \item  The stacked $D_u, D_s$ architecture in SS-GAN where the intermediate features of $D_u$ are fed to $D_s$ ensures that $D_s$, and in turn the generator does not over-fit to the labeled samples. In particular, $D_s$ is forced to learn discriminative features that characterize the association between $x$ and $y$ based on the features over the entire unlabeled set learned by $D_u$, which ensures generalization to the complete set of images.
\end{enumerate}

\bibliographystyle{plain}
\bibliography{references}

\newpage

\newpage

\newpage

\end{document}